# Review on Efficient Strategies for Coordinated Motion and Tracking in Swarm Robotics

B. Udugama, Middlesex University, M00734040

*Abstract*— Swarm robotics is a creative method of organizing multi-robot structures, consisting of many basic robots influenced by communal insects. The greatest astonishing attribute of swarm robots is their capacity to function together to accomplish a collective objective. This paper addresses the list of current surveys, problems and algorithms that were stimulated in the research of Coordinated Movement in Swarm robotics. Algorithms for swarm robotics movement are contrasted, considering the swarm micro-robots to accomplish aggregation, creation, and clamouring by contrasting the relative computational simulations between the algorithms and simulations used.

## I. Introduction

The phrase "Swarm Robotics" refers to advanced group actions that can arise from the mixture of several basic entities, each working independently[1]. Swarm intelligence is, corresponding to Cao et al. [2], "a feature of non-intellectual robots' structures that display collectively sensible behavior." Nevertheless, established on the concepts, that the basic attributes of swarm intellect consist of a naturally motivated focus on autonomous regional regulation and local connectivity and the development of global action because of self-corporation [3]. The adaptation of collaborative robotics of swarm intelligence concepts may be called "Swarm Robotics." Swarm robotics is a novel methodology to integrating large totals of unsophisticated robots [4], which are mobile, not centrally regulated, able to interact locally, and function based on a certain form of biological inspiration. Since the 1980s, swarm robotic techniques have been a significant exploration area [4], as new methods to solutions are being established and tested, the benefits of swarm robotic organisms [2] are also implementable. In 1993, Dudek et al. [8] performed the early effort on classifying exploration areas of swarm robotic techniques. The paper divided the areas into the categories of mapping, biological inspiration, topology of communication, motion coordination, swarm reconfigurability and processing power of swarm units. Cao et al. [2] conducted a hierarchical study of cooperative robotics. They split the publications into community design, resource tensions, collective roots, learning and spatial problems. Through looking at their mutual features,

Luca Iocchi et al. [9] provided an overview of the multirobot networks. They also suggested nomenclature of multi-robot structures and a classification of the multi-robot system's responsive and social purposeful behaviors. Rather of summarizing the swarm robots research field into a classification of cooperative structures [2, 9], Lynne [10] grouped the fields into the key subjects that produced substantial study rates. Responsive study problems within each topic field were also described and specifically addressed in this article.

## II. Biological Inspiration

Swarm engineering and the associated swarm intelligence concept was influenced by an appreciation of the autonomous processes underlying the structure of biological animal behaviors and considering their efficiency. Social insects like ants provide one of the most documented descriptions of self-organized natural actions. They can perform amazing behavioral feats by local and restricted communication: preserving the colony's wellbeing, compassionate for their children, reacting to assault. Thomas et al. [11] studied the actions of a community of robots engaged in an item recovery process in which the control mechanism of the robots is influenced by a foraging behavior pattern of the ants.

The tracks allocated to the automatons are derived from basic ant swarms' activities such as scanning, extracting, depositing, returning and rest. Ideas influenced by these group actions also contributed to the usage of pheromone [10], a biochemical material released by ants and related communal insects to label the area with details for later helping other bees. Likewise, David et al. [8] and Cazangi et al. [7] have used pheromones in their work to accomplish the process of inter-robot contact. More work in this field has culminated in primates being able to communicate and connect with each other.

## III. Mapping

Mapping is a visualization of the actual surroundings through virtual models by sensory data from the mobile robot. Localization is described as evaluating the robot's position within the created spatial structure. The function of the Simultaneous localization and mapping (SLAM) or concurrent mapping and localization (CML) question is to acquire and build a map of an unfamiliar area with the aid of a moving robot when navigating the robot. In the SLAM aspect, due to the situation where the robot entails a global positioning sensor, the robot mostly relies on gradual motion for robot location prediction (e.g., odometry). There are several methods that have been applied to solve the odometry issue in Geometry such as macroscopic mapping and geometric mapping by utilizing different forms of filters. A macroscopic map is an abstract representation of a given ecosystem's structural attributes. For most instances macroscopic maps use points to reflect the world as a series of distinctive positions (e.g., rooms), linked by robot action sequences through lines (e.g. wall-following). Nonetheless, a graphical diagram reflects the environment's exact spatial features, which appears like a floor plan.

## IV. MOTION COORDINATION

Swarm robotics trajectory-planning is a field that has gained a lot of interest over the last two decades. In addition, design a route between two different places for a given robot and an area outline, which must be void of hazards and follow all the optimization criteria; is perceived to be the contemporary problem in the trajectory optimization of the mobile robot. Research of route-planning may be allocated into local and global route-planning to tackle this issue.

Local path determination is constructed using the sensor data provided by transducers mounted on the robot, which provides specifics of the unexplored area. In the other side, the regional preparation determines specifically the layout of the climates, and navigation is carried out with the details established in priori.

## V. MAJOR FACTORS IN SEARCH AND TRACKING

Experts in the history of reasearch discussed various issues when dealing with the topic of target explore and monitoring. These differ in different criteria and expectations, which may limit the study's emphasis to certain sub-complications in effect.

### A. Number of targets

Based on the number of destinations to be sought or monitored, the issue of targeting and monitoring can be split into two major scenarios: one target, and several targets. In order to increase the precision of the objective states calculated, the key priority for tracking a lone objective with any Multi Robot Systems (MRS) is the fusion of sensor data from multiple tracker systems [25].

The situation for multiple goals can be considered by expanding the single goal event, which includes many other complexities. The number of goals, for instance, may be unclear or may also vary over time. However, even though there is information and consistent numbers of targets, sensor observations are still unpredictable because they can come from all the objectives. This is the problem of the correlation of results. And, compared to the single target situation, robots need to disperse to the different objectives needing a job assignment method.

Another crucial element which influences the solution approach is the ratio among the amount of targets and followers [25]. For instance, where the goals are considerably greater than the number of followers it will not always be achievable to track the entire goals and optimize the total number of goals encountered during the project by at least one robot [4]. Instead of monitoring these clusters separately, an alternative method would involve dividing goals into groups [26]. When observing many moving objects, such as a crowd or a bunch of animals, it is not practical or appropriate to observe each person separately. Individuals in a crowd or animals in a flock have the inclination because they can fly more specifically than alone to popular destinations [27,28].

### B. Mobility of targets

The dilemma is either to look for stationary goals or to track moving goals based on the versatility of the goals. While in the Swarm Robotics group the stationary aim case was thoroughly investigated [32], there has been less research contemplating mobile goals [35]. In case of static goals, noisy results, i.e. false alarms, or lack of measurements, are the only confusion. But there is more ambiguity in the goal transition for shifting objectives.

For instance, the goal can travel on the ground, swim underneath water, or rise into the air. The movement modes of the objectives do need to be addressed. Since Swarm Robotics has been a young field of study, most analysis has so far been performed under laboratory conditions in the search and tracking of objects traveling on 2D terrain.

### C. Mobility of followers

Even though followers can mostly be static within wireless sensor frameworks, they are still mobile in the context of a robotic swarm. The tracker function, however, has an significant effect on the solution of the question. This controls the mindset of the followers along with pace and responsiveness. The followers should be the identical as the targets, for instance ground followers which track land moving targets, hovering followers which track airborne targets, etc. The movement approach of the objectives and followers can also vary. Flying followers, for example, may be used to track the flying targets on the ground.

### D. Complexity of environment

Environmental uncertainty is an significant consideration in the creation of the MRS, as robots play a key role in communicating with others and the world. The only relations with the followers and targets to take into consideration in the case of an open room. The atmosphere framework may be used for goal identification or robotics motion preparation in organized settings, such as indoor office-type settings. However, occlusion induced by the ambient configuration should be regarded as inconsistencies in device observations in unorganized configurations, i.e. cluttered ecosystems.

### E. Knowledge of target motion

If mobile targets are tracked via mobile bots, prior experience of goal activity will help you anticipate the next goal position, so that the maneuvers of robots can be regulated respectively. If the objective motion is well understood, it is claimed that the motion pattern is 'deterministic'[25]. The typical case is the usage of rockets to control the projectile as it has a defined course under the rule of physics [37]. In [38], visualization analyses were performed to see how the application of objective complexities in sensor design enhances monitoring efficiency. The discoveries included a community of ground robots which tracked an aerial target with a deterministic motion. Since previous experience of the goal action may be influenced by spontaneous variables [25] the goal motion is considered 'probabilistic.' There seems to be no background knowledge on goal behavior for real-world applications. The alternatives in such situations will either follow a basic motion framework [37,39] or a randomized motion framwork like the Quantum fluctuations [40].

*F. Type of cooperation*

Collaboration between the Swarm Representatives is necessary to obtain the optimal result of the SRS squad, which is supreme to the sheer amount of individual brilliance. The robots can boost their overall efficiency by collaboration in two separate ways:
  (1) the confusion
  (2) allocating the goal.

In the single aim case, the first sort of teamwork is used to integrate observations from several sensors to determine the goal location more precisely than is feasible for a single sensor independently. Throughout the event of objective mission and monitoring through SRS, sensor measurements from various robots may be integrated to approximate the actual target position and pace. They tackled the problem of mutual Multi Robot (MR) monitoring of several mobile objects, based on the integration of sensors [41]. Goal distribution is used in a multi-task system to boost monitoring efficiency by assigning objectives to followers in the right place to track things. It is a Multi-Robot Task Allocation (MRTA) field, wherever the purpose is to delegate tasks to bots in a manner that achieves the broad target more effectively by collaboration [34]. In the issue that the 'tasks' will be independent goals or groups of them, and the objective will be to manage them accurately and effectively.

*G. Coordination among multiple followers*

A strong teamwork approach is mandatory to leverage the full benefits of collaboration between robots. In fact, robot cooperation techniques can be classified into two major groups, specifically explicit cooperation, and implied organization [22]. In specific synchronization, the actions of one bot may be guided by another robot by clear communication [21]. In tacit cooperation, the individual bots create autonomous judgments about how to act, based on the knowledge they obtain from their own experiences and contact with others [24]. By utilizing clear contact, the precision of the knowledge transfer among robots is assured. The contact burden of the device would, however, increase with the total amount of robots, likely degrading device performance [42]. If tacit contact is used, while the knowledge received by the robot is not fully accurate, the overall system's efficiency, responsiveness and fault resilience is improved [42].

VI. MOTION ALGORITHMS FOR SWARMS

As stated in last Segment, the characteristics of the Swarm Robotic Systems (SRS) render themselves quite appropriately equipped for target exploration and monitoring. Within this chapter, we address discover and monitoring algorithms which have been or may possibly be used in SRSs. We group such implementations into two major types: one based on Swarm Intelligent (SI) procedures and the additional based on other methods. Such two types of procedures were discussed independently in the following paragraphs.

*A. Algorithms established on basics of the swarm intelligence*

That's clear to realize the connection among swarm optimization algorithms and Swarm Robotics (SR) exploration algorithms. We also look for 'right places' inside a domain area use swarms. In addition, as Parker[48] states, all main MR engagement abilities, like goal monitoring, require a ruling mechanism that can be articulated as an optimization problem as described out in the below Table 1. It is therefore obvious that Swarm Intelligence procedures can be used to find ideal approaches to tracking algorithms.

Parker[48] also states that such optimization challenges are not commonly regarded as comprehensive optimization problems. The requirement for robots to react in real-time leaves little room to determine globally optimum solutions, except the issue is really small. As a consequence, centralized approaches that integrate only regional cost measures are usually utilized even though they can only estimate the overall solution. Such suboptimal approaches are also appropriate for functional implementations.

Table 1
Optimization of performance challenges

| | Objective | Metric to optimise |
|---|---|---|
| Task allocation [43,33,34] | Map a set of robots to a set of tasks | Optimise overall system utility. Here, "utility" refers to a combination of the quality at which a robot can execute a given task, and the cost it incurs in executing that task (e.g., power consumption) [43]. |
| Path planning [44–47] | Generate paths for multiple robots | Minimise a performance metric e.g., combined robot path lengths [45], combined travel times for robots to reach their respective goals [47], combined energy use [46]. |
| Formations [35] | Enable robots to move into a desired formation, or to maintain a specific formation, while moving through the environment | Minimise the error between each current robot position and that robot's assigned position in the formation. |
| Target tracking or observation [4,35] | Control cooperative robot motions to ensure that a group of targets remains under observation by the robots | Optimise a combination of the time in which targets are under observation and a robot cost function [4]. |

Table 2
Comparison of various goals and analysis

| | Problem Characteristics | | | | | |
|---|---|---|---|---|---|---|
| | Number of targets | Targets/trackers ratio | Mobility of targets | Environment complexity | Prior knowledge of target motion | Cooperation | Coordination |
| Pugh & Martinoli [32] | 1 | ≪ 1 | Stationary | Empty space | N/A | Target estimation | Implicit |
| Parker [4] | Multiple | >1 | Mobile | Empty space | None | Target allocation | Implicit |
| Derr & Manic [31] | Known number | ≤ 1 | Stationary | Cluttered | N/A | Target estimation | Implicit |
| Wang & Gu [38] | 1 | <1 | Mobile | Empty space | None | Target estimation | Implicit |
| Jevtić et al. [34] | Multiple | <1 | Stationary | Cluttered | N/A | Target allocation | Implicit |
| Lee et al. [35] | >1 | ≪ 1 | Mobile | Empty space | None | Target estimation | Implicit |

As SI procedures concentrate primarily on autonomous local regulation, local connectivity, and the development of global action as a consequence of self-organization[7], they obviously match and allow utilization of the main elements of SRs. It can be considered as the primary explanation that a large number of the current SR seek and monitoring algorithms is focused on popular SI procedures.

The scanning and monitoring algorithms outlined in the continuation of this chapter are classified as the initial SI procedures are influenced by. The automated procedures mentioned here have primarily used SI concepts for modeling the actions of particular robots, wherever every single member is viewed as an particle in the accompanying SI procedure.
- Particle swarm optimization
- Bees algorithm
- Artificial Bee Colony Optimisation
- Ant Colony Optimisation
- Bacterial Foraging Optimisation
- Glowworm Swarm Optimisation
- Biased Random Walk
- Firefly Algorithm

*B. Other methods to resolve SI motion optimisation*

This article refers several non Swarm Robotic based methods for aim exploration and monitoring purposes considering the local communication and other parameters.
- Distributed Kalman Filter (DKF)
- Potential fields
- Formation-based target following

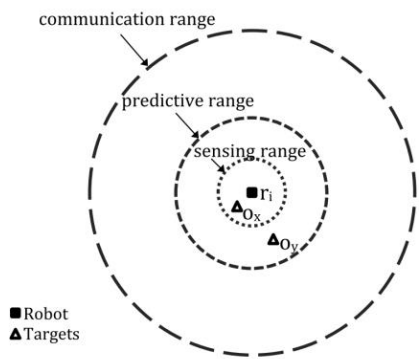

*Fig. 1. Described limits for the isolated Swarm robot particle*

VII. COMPARATIVE ANALYSIS

Goal detection and monitoring issues with and Swarms can be split into two significant smaller challenges. The initial is a goal state prediction for a lone robot, which includes the determination of the goal locations and speeds in the robot's peripheral vision from its instrument observations. The following is the synchronization of movements among robots to monitor further goals across time[25]. Many work in the field of SR focuses on the latter part of the issue.

Table 2 summarizes the different conditions that characterize the problem conditions used in several MR exploration and tracking strategies mentioned in Previous sections. It is apparent as of the desk that only A-CMOMMT [4] tackled the question of monitoring several mobile goals that are greater in quantity than the robot squad.

In order to appreciate the great advantages of the SRS, those assets must also be preserved, as defined in Section 1. The specifications of the different procedures presented in Previous section, with an focus on the characteristics required to match the SRSs are tabulated in Table 3.

With almost no frontrunner, the rigidity of the algorithm enhances as there is no single main fault factor. This also improves the scalability of the algorithm due to decreased coordination between robots. Scalable Performance is the most valuable stuff for every dispersed MRS. It is often valuable to obtain scalability by utilizing only local contact. Demanding global connectivity might not only hinder the distribution of robots due to their restricted scope of contact, but would also trigger overloading of connectivity as the scale of the swarm grows. The absence of robot identifiers also leads to the maintenance of machine scalability. As it is often possible to restrict the amount of single color or graphic pattern indicators to be produced. In addition, identity assignment is a type of centralisation. There is also a limiting presumption of having a common communication method when it comes to SRS. It is difficult for a key inertial navigation approach to track them all because there are very large numbers of robots and the swarm will work at places where GPS and similar schemes are inaccessible [32].

SRS stresses the flexibility and sophistication of different robots when operating in broad quantities. In order to obtain real-time efficiency, something has to be accomplished, with

M00734040 Bavantha Udugama	5Table 3
Swarm Robotics Motion Algorithms feature comparison

|  | Pugh & Martinoli [32] | Derr & Manic [31] | Jevtić et al. [33,34] | Turdeuv et al. [36] | Dhariwal et al. [37] | Parker [4] | Wang & Gu [38] | Lee et al. [35] |
|---|---|---|---|---|---|---|---|---|
| Leaderless | ✓ | ✓ | ✓ | ✓ | ✓ | ✓ | ✓ | ✓ |
| Local/no communication | ✓ |  | ✓ | ✓ |  | ✓ | ✓ | ✓ |
| No robot Identifiers | ✓ |  | ✓ |  | ✓ | ✓ | ✓ | ✓ |
| No common coordinates | ✓ | ✓ |  |  | ✓ |  |  | ✓ |
| Simple computations | ✓ | ✓ | ✓ | ✓ | ✓ | ✓ |  |  |
| No memory of previous states | ✓ |  | ✓ |  |  | ✓ |  | ✓ |
| Mathematically proven properties |  |  |  |  |  |  |  | ✓ |
| Verified through simulations | ✓ | ✓ | ✓ |  | ✓ |  | ✓ | ✓ |
| Verified through real robot experiments |  |  | ✓ | ✓ | ✓ | ✓ | ✓ | ✓ |

ample lightweight computing necessities. SI algorithms appear to be less complicated, so the problem size is also not explicitly connected to the sophisticated system [31]. The calculations per stage in the PSO-based algorithms[32,31] are quite clear and contain just one vector comparison and addition.

In this report, all the SI algorithms mentioned were attempted in simulations in different research papers; and some were also evaluated using real-robot testing[33,37]. The assumption that a significant number of robots can not be used for real-world tests is a temptation to use models in the testing of SRS's. Mostly the works of practical robotics projects only used a substantial number of robots, and used models to check their systems for a larger number of robots. The tests with simulations were also carried out with actual robots to test all non-SI related algorithms [28,4,37,35].

Many aim identification and monitoring systems often involve sampling of the gradients of certain physical, mechanical, genetic or electro magnetic properties to classify possible sources or entities[12]. A successful quest or monitoring algorithm in the gradient sector should not be prone to plateaux (dead space). The inclusion of unpredictability in the algorithm will prevent this issue. The BRW approach mentioned is a good example of it: if the robots assume that the pitch is positive, then they take larger measures in that direction; if a pitch is missing, the robots travel arbitrarily in a set range.

Apart from the more common random Brownian phenomenon, a large variety of species, containing open-ocean rapacious fish[35] and the birds, have also provided significant support for Lévy flight search patterns[36]. Lévy movements are a different form of spontaneous path with the duration of steps from the power-law tail delivery Lévy, rather than the typical Gaussian path used in Brownian travel. This refers to many quick measures of 'rolling fragments' interspersed with larger relocations[36]. Current SI algorithm such as Global Swarm Optimisation (GSO) [37] and also MR quest [30] were adopted for the enhancement of Lévy flights.

The question of several moving targets includes an algorithm that can simultaneously locate many optimal parameters, control Optima, and even manage the usage of proven Optima, when looking for new optima[21]. The procedure will be able to preserve variability in the inhabitants in order to reach several optimum at the same time, such that the whole population is not converging optimally by discovering all optimum[24]. The route of Lévy flights can provide a community diversity toward hasty convergence, effectively exiting the algorithm from local minimum levels[39].

In fact, historical experience was also considered to be valuable for problems of dynamic optimization[25]. Because the present condition of the climate and health is sometimes equivalent to previous ones, it could be simpler to identify viable alternatives in the current world with the use of historical knowledge [12]. In addition, as previous strategies may include additional reference points following a transition, they may lead to adding flexibility to the discovery phase after the quest converges [22].

VIII. CONCLUSIONS

Leading to robustness, versatility, scalability and economic performance, Swarms have the capacity that will be used for real-world activities. Goal identification and tracking are one such activity with a number of applications. In this article, exploration and tracking algorithms for robotic swarms have been identified and compared. The most difficult but exciting scenario of implementation is to use robot swarms to monitor several movement goals among the different issue configurations found in these ventures.

Following many moving goals utilizing a robot swarm is a sophisticated multimodal and vibrant distributed optimization problem. Algorithms for this issue should be able to locate several targets (optima) concurrently, track them, coordinate the manipulation of the established targets and discover new goals. The paper compared several algorithms, including GSO and FA, which give global, yet preliminary solutions to the optimization problem. The dynamics of SI processes are not specifically connected to the problem scale and are thus ideally adapted for SRS implementations in the real world.

REFERENCES

[1] Amanda J. C. Sharkey, "The Application of Swarm Intelligence to Collective Robots" in Advances in Applied Artificial Intelligence, John Fulcher, Idea Group Publishing, 2006, pp. 157 - 185.
[2] Y. Uny Cao, Alex S. Fukunaga and Andrew B. Kahng, "Cooperative Mobile Robotics: Antecedents and Directions", Autonomous Robots, vol. 4, 1997, pp. 1-23.
[3] Amanda J. C. Sharkey, "Swarm Robotics and Minimalism", Connection Science, vol. 19, no. 3, September 2007, pp. 245-260.
[4] Levent Bayindir and Erol Sahin, "A Review of Studies in Swarm Robotics", The Turkish Journal of Electrical Engineering & Computer Sciences, vol. 15, no. 2, 2007, pp. 115-147.

[5] Lynne E. Parker, "Distributed Intelligence: Overview of the Field and its Application in Multi-Robot Systems", Journal of Physical Agents, vol.2, no. 1, March 2008, pp. 5-14.

[6] Altshuler Y., Yanovsky V., Wagner I. A. and Bruckstein A. M., "Swarm Intelligence – Searchers, Cleaners and Hunters" in Swarm Intelligent Systems, Nadia Nedjah and Luiza de Macedo Mourelle, Springer, 2006, pp. 93-132.

[7] Bernardine Dias M. and Anthony (Tony) Stenz, "A Market Approach to Multirobot Coordination", The Robotics Institute, Carnegie Mellon University, Pittsburgh, Pennsylvania 15213, August 2001.

[8] Dudek G., Jenkin M., Millios E. and Wilkes D., "A Taxonomy for Swarm Robots" in Proceedings of the 1993 IEEE/RSJ International Conference on Intelligent Robots and Systems, Yokohama Japan, July 26-30,1993, pp. 441-447.

[9] Luca Iocchi, Daniel Nardi and Massimiliano Salerno, "Reactivity and Deliberation: A Survey on Multi-Robot Systems", Balancing Reactivity and Social Deliberation in Multi-Agent Systems, From RoboCup to Real-World Applications (selected papers from the ECAI 2000 Workshop and additional contributions), Springer-Verlag, London, UK, 2001, pp. 9-34.

[10] Lynne E. Parker, "Current research in Multirobot Systems", Artificial Life Robotics, 2003, pp. 1-5.

[11] Amanda J. C. Sharkey, "Robots, Insects and Swarm Intelligence", Artificial Intelligence Review, vol. 26, 2006, pp. 255-268.

[12] Thomas H. Labella, Marco Dorigo and Jean-Louis Deneubourg, "Division of Labor in a Group of Robots Inspired by Ants' Foraging Behavior", ACM Transactions on Autonomous and Adaptive Systems, vol. 1, no. 1, September 2006, pp. 4-25.

[13] Liviu Panait and Sean Luke, "A Pheromone-Based Utility Model for Collaborative Foraging", in AAMAS'04, New York, USA. : ACM, July 19-23, 2004, pp. 36-43.

[14] David Payton, Regina Estkowski and Mike Howard, "Compound Behaviors in Pheromone Robotics", Robotics and Autonomous Systems, vol.44, 2003, pp. 229–240.

[15] Renato Reder Cazangi, Fernando J. Von Zuben and Maurício F. Figueiredo, "Autonomous Navigation System Applied to Collective Robotics with Ant-Inspired Communication", GECCO'05, Washington, DC, USA. : ACM, June 25-29, 2005, pp. 121-128.

[16] G. Beni, From swarm intelligence to swarm robotics, in: E. Şahin, W.M. Spears (Eds.), Swarm Robotics, in: Lecture Notes in Computer Science, vol. 3342, Springer, Berlin Heidelberg, 2005, pp. 1–9.

[17] G. Beni, J. Wang, Swarm intelligence in cellular robotic systems, in: Proceed. NATO Advanced Workshop on Robots and Biological Systems, 1989.

[18] T. Fukuda, S. Nakagawa, Y. Kawauchi, M. Buss, Self organizing robots based on cell structures—cebot, in: IEEE International Workshop on Intelligent Robots, 1988, 1988, pp. 145–150.

[19] T. Fukuda, Y. Kawauchi, Cellular robotic system (cebot) as one of the realization of self-organizing intelligent universal manipulator, in: 1990 IEEE International Conference on Robotics and Automation, 1990, Proceedings, vol.1, 1990, pp. 662–667.

[20] A.J. Sharkey, N. Sharkey, The application of swarm intelligence to collective robots, in: J. Fulcher (Ed.), Advances in Applied Artificial Intelligence, IGI Global, 2006, pp. 157–185.

[21] M. Shimizu, A. Ishiguro, A self-reconfigurable robotic system that exhibits amoebic locomotion, in: IEEE/ICME International Conference on Complex Medical Engineering, 2007. CME 2007, 2007, pp. 101–106.

[22] M. Shimizu, A. Ishiguro, An amoeboid modular robot that exhibits real- time adaptive reconfiguration, in: Proceedings of the 2009 IEEE/RSJ International Conference on Intelligent Robots and Systems, IROS'09, IEEE Press, Piscataway, NJ, USA, 2009, pp. 1496–1501.

[23] H. Chung, S. Oh, D. Shim, S. Sastry, Toward robotic sensor webs: Algorithms, systems, and experiments, Proc. IEEE 99 (9) (2011) 1562–1586.

[24] M. Yogeswaran, S.G. Ponnambalam, Swarm Robotics: An Extensive Research Review, in: I. Fuerstner (Ed.), Advanced Knowledge Application in Practice, InTech, 2010, Ch. 14.

[25] B. Jung, Cooperative target tracking using mobile robots, Ph.D. thesis University of Southern California, Los Angeles, CA, USA, 2005, aAI3180426.

[26] F. Armaghani, I. Gondal, J. Kamruzzaman, D. Green, Dynamic clusters graph for detecting moving targets using wsns, in: 2012 IEEE Vehicular Technology Conference (VTC Fall), 2012, pp. 1–5.

[27] J.J. Faria, E.A. Codling, J.R. Dyer, F. Trillmich, J. Krause, Navigation in human crowds; testing the many-wrongs principle, Anim. Behav. 78 (3) (2009) 587–591.

[28] J. Krause, G.D. Ruxton, S. Krause, Swarm intelligence in animals and humans, Trends Ecol. Evol. 25 (1) (2010) 28–34.

[29] C. Li, S. Yang, A clustering particle swarm optimizer for dynamic optimization, in: IEEE Congress on Evolutionary Computation, 2009. CEC'09, 2009, pp. 439–446.

[30] S. Yang, C. Li, A clustering particle swarm optimizer for locating and tracking multiple optima in dynamic environments, IEEE Trans. Evol. Comput. 14 (6) (2010) 959–974.

[31] K. Derr, M. Manic, Multi-robot, multi-target particle swarm optimization search in noisy wireless environments, in: 2nd Conference on Human System Interactions, 2009. HSI'09, 2009, pp. 81–86.

[32] J. Pugh, A. Martinoli, Inspiring and modeling multi-robot search with particle swarm optimization, in: Swarm Intelligence Symposium, 2007, SIS 2007, IEEE, 2007, pp. 332–339.

[33] A. Jevtić, P. Gazi, D. Andina, M. Jamshidi, Building a swarm of robotic bees, in: World Automation Congress (WAC), 2010, 2010, pp. 1–6.

[34] A. Jevtić, A. Gutierrez, D. Andina, M. Jamshidi, Distributed bees algorithm for task allocation in swarm of robots, IEEE Syst. J. 6 (2) (2012) 296–304.

[35] G. Lee, N. Chong, H. Christensen, Tracking multiple moving targets with swarms of mobile robots, Intell. Serv. Robot. 3 (2010) 61–72.

[36] K. Krishnanand, D. Ghose, Chasing multiple mobile signal sources: A glowworm swarm optimization approach, in: Proceedings of the 3rd Indian International Conference on Artificial Intelligence (IICAI-07), Pune, India, 2007, pp. 1308–1327.

[37] T. Kirubarajan, Y. Bar-Shalom, Y. Wang, Passive ranging of a low observable ballistic missile in a gravitational field, IEEE Trans. Aerosp. Electron. Syst. 37 (2) (2001) 481–494.

[38] J. Spletzer, C. Taylor, Dynamic sensor planning and control for optimally tracking targets, Int. J. Robot. Res. 22 (1) (2003) 7–20.

[39] B. Jung, G.S. Sukhatme, Detecting moving objects using a single camera on a mobile robot in an outdoor environment, in: International Conference on Intelligent Autonomous Systems, The Netherlands, 2004, pp. 980–987.

[40] D. Montemerlo, S. Thrun, W. Whittaker, Conditional particle filters for simultaneous mobile robot localization and people-tracking, in: IEEE International Conference on Robotics and Automation, 2002. Proceedings. ICRA'02, vol. 1, 2002, pp. 695–701.

[41] M. Powers, R. Ravichandran, F. Dellaert, T. Balch, Improving multirobot multitarget tracking by communicating negative information, in: L. Parker,

[42] F. Schneider, A. Schultz (Eds.), Multi-Robot Systems, in: From Swarms to Intelligent Automata, vol. III, Springer, Netherlands, 2005, pp. 107–117.

[43] Z. Yan, N. Jouandeau, A. Ali Cherif, A survey and analysis of multi-robot coordination, Int. J. Adv. Robot. Syst. 10 (399) (2013).

[44] B.P. Gerkey, M.J. Matarić, A formal analysis and taxonomy of task allocation in multi-robot systems, Int. J. Robot. Res. 23 (9) (2004) 939–954.

[45] L.E. Parker, Multiple mobile robot teams, path planning and motion coordination, in: R.A. Meyers (Ed.), Encyclopedia of Complexity and Systems Science, Springer, New York, 2009, pp. 5783–5800.

[46] A. Banharnsakun, T. Achalakul, R. Batra, Target finding and obstacle avoidance algorithm for microrobot swarms, in: 2012 IEEE International Conference on Systems, Man, and Cybernetics (SMC), 2012, pp. 1610–1615.

[47] D. Michel, K. McIsaac, New path planning scheme for complete coverage of mapped areas by single and multiple robots, in: 2012 International Conference on Mechatronics and Automation (ICMA), 2012, pp. 1233–1240.

[48] M. Kapanoglu, M. Alikalfa, M. Ozkan, A. Yazıcı, O. Parlaktuna, A pattern- based genetic algorithm for multi-robot coverage path planning minimizing completion time, J. Intell. Manuf. 23 (4) (2012) 1035–1045.